\documentclass{article} 
\usepackage{aditi}
\renewcommand{\ForumContactRow}{%
  \begingroup\small\raggedright
    \ifx\ForumEmail\empty\else
      {\color{ForumAccent}\faEnvelope[regular]~}\ %
      \href{mailto:\ForumEmail}{\textcolor{ForumContactText}{\texttt{\ForumEmail}}}\par
      \vspace{\ForumContactGap}%
    \fi
    
  \endgroup
}
\usepackage{microtype}
\usepackage{xspace}
\usepackage{hyperref}
\usepackage{url}
\usepackage{booktabs}

\usepackage{style} 
\definecolor{skyblue}{RGB}{204,229,255}

\usepackage{algorithm}
\usepackage{algorithmic}

\usepackage[most]{tcolorbox}
\tcbset{
  lightbluebox/.style={
    colback=skyblue!70,        
    colframe=skyblue!75!black, 
    boxrule=1pt,
    arc=4pt,
    auto outer arc
  }
}

\definecolor{darkblue}{rgb}{0, 0, 0.5}
\hypersetup{colorlinks=true, citecolor=darkblue, linkcolor=darkblue, urlcolor=darkblue}

\setTitleruleGap{0.75pt}         

\title{Harmonizing Dense and Sparse Signals in Multi-turn RL: Dual-Horizon Credit Assignment for Industrial Sales Agents}

\setauthors{Haojin Yang$^{1}$ \authorsep Ai Jian$^{2}$\ \authorsep Xinyue Huang$^{3}$  \authorsep Yiwei Wang$^{4}$ \authorsep Weipeng Zhang$^{5}$ \authorsep Ke Zeng$^{5}$ \authorsep Xunliang Cai$^{5}$ \authorsep Jingqing Ruan$^{5}$\thanks{Corresponding author.}}
\setaffils{$^{1}$Peking University,\quad $^{2}$Beijing University of Posts and Telecommunications,\quad $^{3}$Sun Yat-sen University,\quad $^{4}$University of California at Merced,\quad $^{5}$Meituan }

\setemail{yhj@stu.pku.edu.cn, ruanjingqing@meituan.com}

\usepackage{xcolor}
\usepackage{minitoc}
\usepackage{graphicx}
\definecolor{jcg}{RGB}{100,160,0}
\definecolor{sachin}{RGB}{0,0,150}
\definecolor{hqz}{RGB}{160,100,100}
\definecolor{gnz}{HTML}{64B5F6}
\usepackage[most]{tcolorbox} 
\tcbuselibrary{skins,breakable} 

\definecolor{myDarkGreen}{RGB}{50, 70, 70} 
\definecolor{myLightGray}{RGB}{240, 240, 240} 

\definecolor{titlebgcolor}{RGB}{70, 80, 100}
\definecolor{bodybgcolor}{RGB}{245, 245, 245}
\definecolor{bordercolor}{RGB}{120, 120, 120}
\definecolor{darkblue}{rgb}{0.0, 0.0, 0.55}   
\definecolor{darkgreen}{rgb}{0.0, 0.5, 0.0}   
\definecolor{darkred}{rgb}{0.6, 0.0, 0.0}     

\usepackage{url}
\usepackage{amsthm}
\usepackage{minitoc}
\usepackage{enumitem}
\usepackage{todonotes}
\usepackage{float}
\usepackage{listings}
\usepackage{caption}
\usepackage[table]{xcolor} 
\usepackage{graphicx}

\usepackage{booktabs}

\definecolor{myLightBlue}{RGB}{230, 240, 255} 

\usepackage[T1]{fontenc}

\newtcolorbox{responsebox}[2][]{
    breakable,
    enhanced,
    colback=white,             
    colframe=blue!50!black,    
    coltext=black,             
    coltitle=white,    
    fonttitle=\bfseries\rmfamily, 
    arc=3mm,                   
    boxrule=1pt,
    title=#2,
    #1
}

\definecolor{lightblue}{RGB}{235,243,252}

\definecolor{mybgcolor}{RGB}{235, 235, 250}
\definecolor{myGreen}{RGB}{240, 250, 240}

\newtcolorbox{takeawaybox}[1][]{
  enhanced,
  colback=mybgcolor, 
  colframe=black,    
  boxrule=0.5pt,     
  arc=3mm,           

  attach boxed title to top left={yshift=-0.25em, xshift=1em},
  fonttitle=\bfseries, 
  title={#1},          
  boxed title style={
    colback=black,     
    sharp corners,     
  },
}
\newtcolorbox{equationbox}[1]{
  colback=white,                
  colframe=gray!75!black,       
  boxrule=1pt,                  
  
  title=#1,                     
  attach boxed title to top left={yoffset=-2mm, xshift=2mm}, 
  
  colbacktitle=gray!75!black,   
  coltitle=white,               
  fonttitle=\bfseries\sffamily, 
  
  boxed title style={
    boxrule=0pt,                
    frame code={}               
  }
}

\begin{document}

\maketitle

\begin{abstract}
Optimizing large language models for industrial sales requires balancing long-term commercial objectives (e.g., conversion rate) with immediate linguistic constraints such as fluency and compliance. Conventional reinforcement learning often merges these heterogeneous goals into a single reward, causing high-magnitude session-level rewards to overwhelm subtler turn-level signals, which leads to unstable training or reward hacking.
To address this issue, we propose \textbf{Dual-Horizon Credit Assignment (DuCA)}, a framework that disentangles optimization across time scales. Its core, \textbf{Horizon-Independent Advantage Normalization (HIAN)}, separately normalizes advantages from turn-level and session-level rewards before fusion, ensuring balanced gradient contributions from both immediate and long-term objectives to the policy update.
Extensive experiments with a high-fidelity user simulator show DuCA outperforms the state-of-the-art GRPO baseline, achieving a $6.82\%$ relative improvement in conversion rate, reducing inter-sentence repetition by $82.28\%$, and lowering identity detection rate by $27.35\%$, indicating a substantial improvement for an industrial sales scenario that effectively balances the dual demands of strategic performance and naturalistic language generation.
\end{abstract}

 \section{Introduction}

\begin{figure}[!ht]
  \includegraphics[width=\columnwidth]{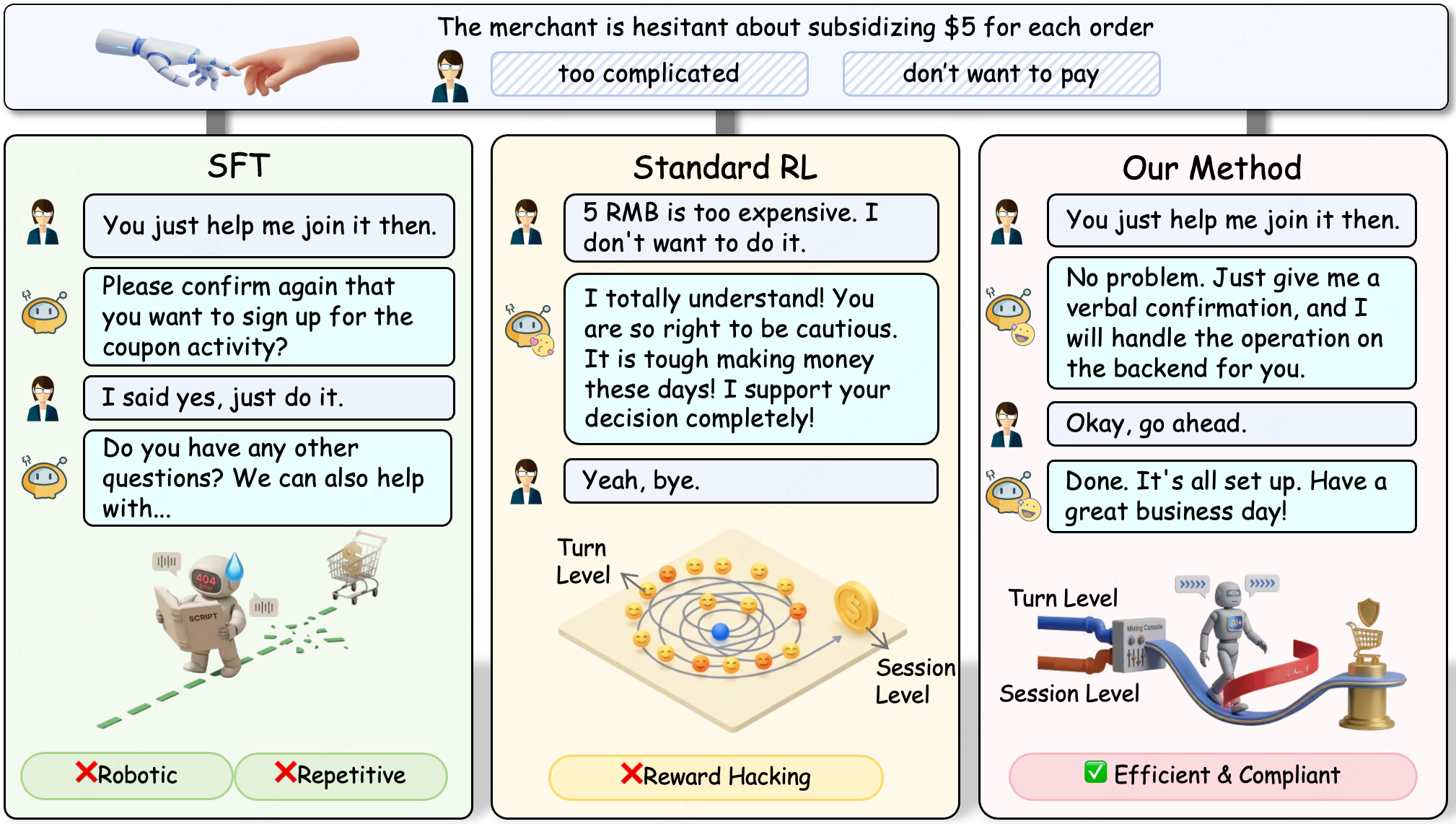}
  \caption {Comparison of dialogue strategies between SFT, Standard RL, and our proposed DuCA method.}
  \label{fig: background}
\end{figure}


Large language models have demonstrated remarkable capabilities in open-domain conversation~\citep{cheng2026highersatisfactionlowercost, feng2025doctoragentrlmultiagentcollaborativereinforcement} and instruction following~\citep{Guo_2025} tasks. 
However, deploying LLMs for high-stakes industrial applications, such as professional sales, remains a significant challenge.
Unlike generic assistants, sales dialogues are inherently long-horizon and strictly goal-driven.
An effective sales agent must not only maintain conversational fluency but also strategically guide interactions toward conversion, all while adhering to stringent compliance constraints~\citep{dong-etal-2025-protod}.
While supervised fine-tuning~\citep{ouyang2022traininglanguagemodelsfollow} can impart stylistic nuances and tone, it often fails to capture the long-term planning required to maximize conversion rates across multi-turn exchanges.

Reinforcement learning (RL) offers a promising avenue for optimizing such goal-oriented policies~\citep{shao2024deepseekmathpushinglimitsmathematical}. Yet, applying RL in sales dialogues exposes a fundamental temporal credit assignment dilemma.
The primary business objectives (conversion rate) are sparse, session-level signals that are only revealed at the end of a conversation, while linguistic quality and engagement are dense, turn-level signals. 
Simultaneous optimization of these conflicting objectives is notoriously unstable. Naive reward aggregation frequently suffers from gradient dominance: high-magnitude, sparse rewards can overshadow nuanced conversational skills, whereas dense, local signals may encourage reward hacking, causing the agent to prioritize short-term gains over ultimate conversion.


To address these challenges, we propose a robust multi-turn RL framework tailored for industrial sales agents. We construct a high-fidelity user simulator to facilitate extensive multi-turn interactions, mitigating the risks and costs of online exploration. 
At its core is Dual-Horizon Credit Assignment (DuCA), which employs Horizon-Independent Advantage Normalization (HIAN) to disentangle optimization across time scales.
Unlike traditional scalarization methods, DuCA treats turn-level guidance (e.g., linguistic heuristics) and session-level objectives (e.g., conversion and compliance) as distinct supervision signals. We normalize advantages independently for each granularity before fusion, ensuring balanced policy updates that integrate both immediate interaction patterns and long-term strategic incentives. As exemplified in Fig\ref{fig: background}, unlike SFT and standard RL which often lead to robotic or sycophantic responses, DuCA achieves a more strategic and compliant dialogue flow by effectively balancing these dual-horizon objectives.
Our contributions are summarized as follows:
\begin{itemize}
    \item \textbf{Industrial multi-turn training framework}: We propose a comprehensive training system for multi-turn dialogues, designing a high-fidelity user simulator to provide high-quality, interactive conversational data, which facilitates extensive multi-turn policy exploration and bridges the gap between static supervised learning and dynamic real-world deployment.
    \item \textbf{Dual-horizon credit assignment mechanism}: We introduce the horizon-independent advantage normalization, which independently normalizes advantages from turn-level and session-level rewards, effectively addressing optimization instability caused by multi-scale reward signals.
    \item \textbf{Empirical effectiveness}: Extensive experiments demonstrate that our approach substantially outperforms standard SOTA RL baselines, achieving a $6.82\%$ relative improvement in conversion rate, reducing inter-sentence repetition by $82.28\%$, and lowering identity detection rate by $27.35\%$, validating its practical value for large-scale industrial deployment.
\end{itemize}
 \section{Related Work}
\paragraph{Multi-turn RL for dialogue.} Recent advancements in aligning LLMs for multi-turn interactions have moved beyond simple SFT. We categorize existing reinforcement learning (RL) approaches into two main streams: \textbf{Process-based Credit Assignment:} Traditional methods typically rely on Process Reward Models (PRMs) or critic models to assign credit to intermediate states \citep{schulman2017proximalpolicyoptimizationalgorithms, jian2026patarmbridgingpairwisepointwise}. However, these entail significant training overhead and heavy reliance on PRM quality. To eliminate critic dependency, recent studies like TARL \citep{tan2025processsupervisedreinforcementlearninginteractive} and GiGPO \citep{feng2025groupingrouppolicyoptimizationllm}  adapt preference optimization to multi-turn contexts using fine-grained rules or state clustering. A critical limitation of these methods is the assumption that step-level rewards align consistently with the final outcome. In industrial sales, immediate linguistic constraints (e.g., compliance) often conflict with aggressive long-term objectives (e.g., conversion), rendering simple dense reward integration insufficient.


\paragraph{Credit assignment in Multi-turn RL.} Effective credit assignment is pivotal for learning from mixed signals in multi-turn training. Existing research primarily explores reward granularity and fusion architectures: MT-GRPO \citep{wei2025reinforcingmultiturnreasoningllm} propagates outcome rewards backward using GAE  \citep{schulman2018highdimensionalcontinuouscontrolusing}, while MGR  \citep{anonymous2026both}  employs turn-level rewards as a gating mechanism. PURE \citep{cheng2025stopsummationminformcredit}  adopts a conservative approach by minimizing future rewards to penalize worst-case outcomes. Despite these strategies, most prior works merge different reward signals into one value, usually by summing or using gating mechanisms, and then normalize the result \citep{yang2025treerpotreerelativepolicy}. This coupling leads to gradient dominance, where high-variance trajectory signals overwhelm subtle turn-level signals. Unlike the aforementioned works, our DuCA framework decouples these horizons during the \textit{normalization} phase. This prevent strategic goals from being diluted by dense interaction constraints, achieving a robust balance that avoids the passive behaviors typical of conservative frameworks like PURE.
 \section{Method}
\label{sec:method}
\begin{figure*}[t]
  \centering
  \includegraphics[width=\linewidth]{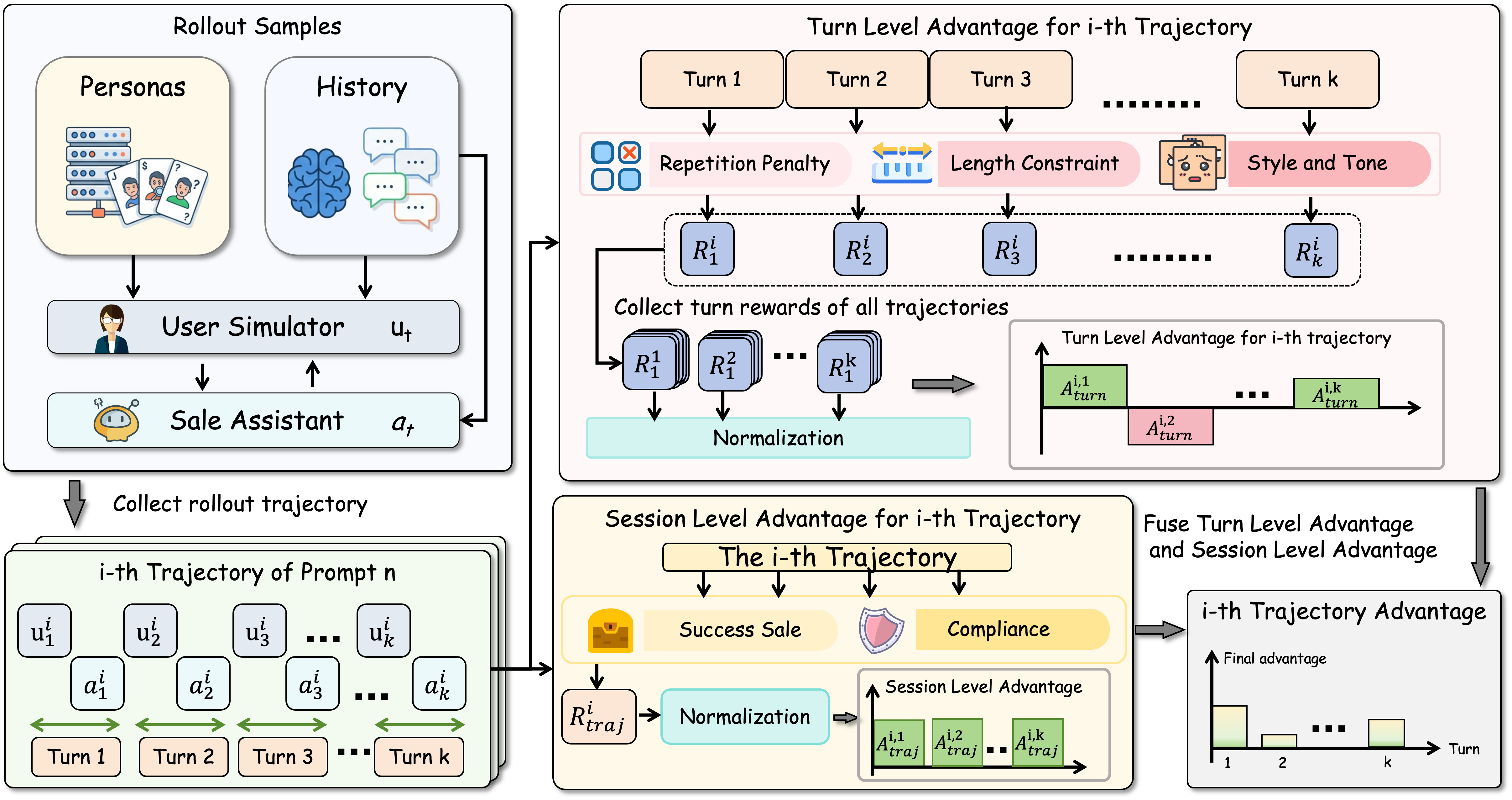}
  \caption {Overview of the DuCA framework. The system generates interaction trajectories via a user simulator conditioned on personas and history. It independently calculates and normalizes: (1) Turn-level advantages from dense heuristic constraints, and (2) Session-level advantages from sparse business outcomes. These decoupled signals are fused to provide a balanced final advantage for robust policy updates.}
    \label{fig:main_framework}
\end{figure*}


In this section, we introduce \textbf{Du}al-Horizon \textbf{C}redit \textbf{A}ssignment mechanism (\textbf{DuCA}), a robust multi-turn RL framework for sales agents in Figure~\ref{fig:main_framework}. 
We elaborated the problem formulation, high-fidelity user simulator, and our core contribution: \textbf{Horizon-Independent Advantage Normalization (HIAN)} for balancing dense and sparse signals.



\subsection{Problem Formulation}
We formulate the multi-turn sales dialogue as a Markov Decision Process, $\mathcal{M} = (\mathcal{S}, \mathcal{A}, \pi, \mathcal{R}, \gamma)$. At each turn $t$, the state $s_t \in \mathcal{S}$ encompasses the full dialogue history $s_t = \{u_1, a_1, \dots, u_t\}$, where $u_t$ is the user query generated by a simulator $\pi_{sim}(u_t|s_{t-1}, \psi)$ with the persona $\psi$ and $a_t \sim \pi_\theta(a_t|s_t)$ is the agent's response. The reward $R(s_t, a_t)$ is decomposed into dense turn-level rewards $r_{turn}^{(t)}$ for linguistic quality and a sparse session-level reward $R_{session}$ for final conversion and compliance. Our objective is to maximize the expected multi-granularity reward:
{\small
\begin{equation}
    \mathcal{J}(\theta) = \mathbb{E}_{\tau \sim (\pi_\theta, \pi_{sim})} \left[ \sum_{t=0}^{T-1} \gamma^t r_{turn}^{(t)} + \gamma^T R_{session}(\tau) \right],
\end{equation}}
where $\tau$ represents a trajectory influenced by the strategic interaction between the agent and the personalized user simulator.

\subsection{Environment: High-fidelity User Simulator with Personas}

Direct training via interaction with real customers entails high operational risks and suffers from sparse feedback. To address this, we construct a high-fidelity user simulator $\mathcal{E}$ based on a powerful LLM, serving as a proxy for real-world customers.

We formulate the user simulator as a conditional generative policy $\pi_{sim}(u_t \mid s_t, \psi)$, which generates a user utterance $u_t$ at turn $t$. This process is conditioned on three key components:

\begin{itemize}[itemsep=0pt, topsep=0pt, leftmargin=15pt]
    \item \textbf{Dialogue context ($s_t$)}: The interaction history $s_t = \{u_1, a_1, \dots, u_{t-1}, a_{t-1}\}$, representing shared grounding and information state between the agent and the simulator.
    \item \textbf{User Persona ($\psi$)}: A static attribute set $\psi \in \Psi$ extracted for each episode (e.g., \textit{Price-sensitive}, \textit{skeptical}), which conditions the simulator's linguistic style and decision logic to mimic diverse real-world customer behaviors.
\end{itemize}

This framework enables $\mathcal{E}$ to mimic complex, non-stationary customer decision-making behaviors, effectively bridging the simulation-to-reality gap for policy exploration.

\subsection{Multi-Granularity Reward Design}
To address the dual objectives of sales dialogues, we decompose the reward signal $R$ into two distinct granularities: dense \textbf{turn-level rewards} and sparse \textbf{session-level rewards}.

\subsubsection{Turn-level Rewards ($r_{\text{turn}}$)}
Turn-level rewards provide immediate and dense feedback, aimed at guiding fundamental conversational skills and maintaining dialogue consistency. Given the dense and immediate nature of turn-level interactions, we employ predefined heuristic rules to model these intermediate rewards. This approach provides stable and explicit value signals, thereby stabilizing the training trajectory. We define $r_{\text{turn}}(s_t, a_t)$ based on the following heuristic rules and utility functions: repetition penalty, length constraint, and style and tone, elaborated in the Appendix~\ref{app:turn_level}.

For each turn $t$, these sub-rewards are aggregated via a gating mechanism (e.g., penalties override style rewards) to form the final scalar $r_{\text{turn}}^{(t)}$.





\subsubsection{Session-level Rewards ($R_{\text{session}}$)}
Session-level rewards reflect ultimate business objectives and strictly enforced safety constraints. These signals are sparse and typically determined only at the end of the dialogue (step $T$). We define the session reward for a completed episode $\tau$ as:
$$
    R_{\text{session}}(\tau) = \alpha \cdot \mathbb{I}(\text{Conversion}) + \beta \cdot S_{\text{compliance}}(\tau),
$$
where $\mathbb{I}(\cdot)$ is an indicator function for successful conversion, and $S_{\text{compliance}}$ is a scoring function that penalizes regulatory violations, detailed in the Appendix~\ref{app:session_level}.

\subsection{Dual-Horizon Credit Assignment (DuCA)}
\label{sec:duca}

A naive scalarization of rewards (i.e., simply summing $r_t^{\text{total}} = r_{\text{turn}}^{(t)} + r_{\text{session}}^{(t)}$) often leads to optimization instability. This is primarily due to the \textbf{gradient dominance} problem: high-magnitude, high-variance sparse rewards (e.g., a large bonus for a successful sale) can overwhelm subtle dense signals, or conversely, dense rewards can induce reward hacking where the agent ignores long-term goals. To resolve this, we propose the \textbf{DuCA} mechanism, which disentangles the credit assignment process into three strategic steps.

\paragraph{Step 1: Independent Advantage Estimation.}
We maintain two separate value heads, $V_{\phi_{\text{turn}}}$ and $V_{\phi_{\text{session}}}$, to estimate expected returns for turn-level and session-level objectives respectively. We compute the Generalized Advantage Estimation (GAE) separately:
\begin{align}
    A_{\text{turn}}^{(t)} &= \text{GAE}(r_{\text{turn}}, V_{\phi_{\text{turn}}}, \lambda_{turn}, \gamma_{turn}), \\
    A_{\text{session}}^{(t)} &= \text{GAE}(r_{\text{session}}.V_{\phi_{\text{session}}}, \lambda_{session}, \gamma_{session}),
\end{align}
This decoupling enables tailored temporal dynamics. We employ standard GAE parameters ($\gamma_{\text{turn}} = 0.99, \lambda_{\text{turn}} = 0.95$) for local fluency to balance bias and variance. Conversely, we set $\gamma_{\text{session}} = \lambda_{\text{session}} = 1.0$, ensuring the sparse terminal reward is propagated uniformly to all actions without decay, effectively bridging the long-horizon gap.



\paragraph{Step 2: Horizon-Independent Advantage Normalization (HIAN).}
To ensure balanced gradient contributions, we normalize advantages independently within each mini-batch $\mathcal{B}$:
\begin{align}
    \hat{A}_{\text{turn}} &= \frac{A_{\text{turn}} - \mu_{\mathcal{B}}(A_{\text{turn}})}{\sigma_{\mathcal{B}}(A_{\text{turn}}) + \epsilon}, \\
    \hat{A}_{\text{session}} &= \frac{A_{\text{session}} - \mu_{\mathcal{B}}(A_{\text{session}})}{\sigma_{\mathcal{B}}(A_{\text{session}}) + \epsilon}.
\end{align}


\paragraph{Theoretical Justification.}
We analyze why HIAN outperforms naive scalarization. In naive methods, the effective gradient is scaled by the total variance: $\Delta \theta \propto (A_{\text{turn}} + A_{\text{session}}) / \sqrt{\sigma_{\text{turn}}^2 + \sigma_{\text{session}}^2}$. Since sparse business rewards typically have high variance ($\sigma_{\text{session}} \gg \sigma_{\text{turn}}$), the contribution of the turn-level signal becomes $\approx A_{\text{turn}} / \sigma_{\text{session}} \to 0$. This \textbf{gradient suppression} halts the learning of linguistic skills. HIAN decouples this dependency, scaling $A_{\text{turn}}$ by its own deviation $\sigma_{\text{turn}}$, ensuring robust learning of both objectives simultaneously.

\paragraph{Step 3: Strategic Fusion and Optimization.}
The final advantage is a weighted combination: $A_{\text{total}}^{(t)} = w_{\text{turn}} \hat{A}_{\text{turn}}^{(t)} + w_{\text{session}} \hat{A}_{\text{session}}^{(t)}$. The policy $\pi_\theta$ is updated via the PPO objective:
\begin{equation}
\begin{split}
\mathcal{L}(\theta) = & \mathbb{E}_t \Big[ \min \big( \rho_t(\theta) A_{\text{total}}^{(t)}, \\
& \text{clip}(\rho_t(\theta), 1-\epsilon, 1+\epsilon) A_{\text{total}}^{(t)} \big) \Big],
\end{split}
\end{equation}
where $\rho_t(\theta) = \frac{\pi_\theta(a_t|s_t)}{\pi_{\theta_{old}}(a_t|s_t)}$ is the probability ratio. This ensures strategic planning without sacrificing conversational quality.


 
\section{Experiments}
\label{sec:experiments}

\begin{table*}[h]
\centering
\caption{Main Results on Sales Dialogue Evaluation. CVR and Compliance represent primary business outcomes, while the remaining metrics evaluate fine-grained interaction quality. Best results across all models are \textbf{bolded}.}
\label{tab:main_results}
\resizebox{\linewidth}{!}{
\begin{tabular}{l|cc|ccccccc}
\hline
\textbf{Method} & \textbf{CVR($\uparrow$)} & \textbf{Compliance($\uparrow$)} & \textbf{Avg. Turn} & \textbf{Intra-R($\downarrow$)} & \textbf{Inter-R($\downarrow$)} & \textbf{IDR($\downarrow$)} & \textbf{Prefix-R($\downarrow$)} & \textbf{Filler($\downarrow$)} & \textbf{PATR($\uparrow$)} \\ \hline
\textit{Foundation Models} & & & & & & & & & \\
DeepSeek-R1 & 22.85\% & 65.13 & 9.85 & 35.19\% & \textbf{1.01\%} & \textbf{2.50\%} & \textbf{7.69\%} & 59.82\% & 42.74\% \\
Longcat-Flash-Chat & 19.99\% & 70.17 & 9.27 & 3.02\% & 23.13\% & 5.92\% & 39.10\% & \textbf{33.35\%} & 43.65\% \\ \hline
\textit{Training Methods} & & & & & & & & & \\
SFT (Base) & 22.23\% & 59.11 & 11.41 & 9.37\% & 35.87\% & 10.59\% & 52.30\% & 51.70\% & 43.59\% \\
REINFORCE++ & 21.51\% & 61.32 & 11.18 & 8.34\% & 30.17\% & 9.99\% & 50.34\% & 54.69\% & 43.62\% \\
GRPO & 22.88\% & 65.40 & 12.49 & 5.48\% & 15.29\% & 9.28\% & 47.76\% & 61.19\% & 43.30\% \\
GDPO & 22.51\% & \textbf{68.53} & 12.19 & 6.48\% & 17.22\% & 8.41\% & 50.03\% & 55.96\% & 44.11\% \\
\textbf{DuCA (Ours)} & \textbf{24.44\%} & 66.72 & \textbf{9.91} & \textbf{4.20\%} & 2.71\% & 6.11\% & 34.44\% & \textbf{49.21\%} & \textbf{44.61\%} \\ \hline
\end{tabular}
}
\end{table*}

\subsection{Experimental Setup}

\noindent\paragraph{Datasets and simulator.} We constructed our training dataset based on 31,000 anonymized real-world business sales dialogues. Additionally, we captured 10,000 high-quality online samples to fine-tune an internal LLM-based user simulator, serving as a high-fidelity interactive environment. The simulator uses heterogeneous user personas (e.g., price-sensitive, skeptical) to mimic complex real-world decision-making.

\noindent\paragraph{Implementation Details and Baselines.} Our sales agent policy $\pi_\theta$ is initialized from a proprietary model after supervised fine-tuning. We evaluate DuCA against four RL baselines initialized from the same base model: \textbf{SFT} (behavioral cloning lower bound); \textbf{REINFORCE++} (linear reward scalarization); \textbf{GRPO} (group-wise advantage normalization); and \textbf{GDPO} (independent normalization for heterogeneous sources). Training is optimized via PPO with a KL-divergence penalty.

\noindent\paragraph{Metrics and Protocol.} We focus on \textbf{Conversion Rate (CVR)} for business outcome and \textbf{Compliance Score} for regulatory adherence. Evaluations use an LLM-as-a-Judge paradigm on generated multi-turn trajectories from a held-out test set.

\subsection{Main Results}

Table \ref{tab:main_results} summarizes the performance across primary business outcomes and fine-grained interaction quality metrics. Overall, our DuCA method significantly outperforms all baseline algorithms, demonstrating a superior Pareto balance between conversion and compliance.

While foundation models like Longcat-Flash-Chat and DeepSeek-R1 exhibit high compliance (up to 77.23\%) and stylistic diversity, they often lack the domain-specific strategic depth required for industrial sales. DeepSeek-R1's CVR (22.85\%) is surpassed by DuCA, indicating that standard models, while fluent, are not fully optimized for goal-oriented industrial conversion.

DuCA achieves a 24.44\% CVR, which is a 6.82\% relative improvement over the strongest baseline GRPO, showing the superiority of our mwthod in exploring effective sales strategies. Additionally, the 7.61\% improvement in compliance over SFT demonstrates that our framework effectively maintains safety boundaries while pursuing aggressive conversion goals.

Our method achieves a 2.71\% Inter-turn Repetition rate, representing an 82.28\% relative reduction compared to GRPO, which proves that the dual-horizon assignment effectively solves the "repetition loop" common in standard RL. Furthermore, the reduction of the Identity Detection Rate by 27.35\% relative to GDPO indicates that DuCA produces more professional interactions.

DuCA reduces the Average Turn by 11.36\% relative to REINFORCE++, while increasing the Positive Attitude Transfer Rate (PATR) to 44.61\%, illustrating the superiority of our method in steering user intent toward conversion with higher operational efficiency.

These consistent gains across heterogeneous metrics suggest that the bottleneck in sales-oriented RL is the gradient interference between dense linguistic rewards and sparse business signals. By decoupling these horizons via HIAN, DuCA successfully mitigates gradient dominance, ensuring that turn-level constraints are not overwhelmed by high-variance session outcomes. This allows the agent to learn a balanced policy that is both persuasively strategic and linguistically compliant.

In summary, the experimental evidence confirms that DuCA establishes a new state-of-the-art for industrial sales agents, delivering a 6.82\% CVR boost and a dramatic 82.28\% drop in conversational redundancy. These results validate that hierarchical credit assignment is a robust and scalable paradigm for aligning autonomous agents with complex, multi-granular industrial constraints.

\subsection{Ablation Study}

To verify the contribution of each core component, we designed the following variants:
\begin{itemize}
    \item \textbf{w/o HIAN}: Removes the decoupled normalization strategy and uses standard reward summation for advantage calculation.
    \item \textbf{w/o Multi-turn}: Trains on single-turn interaction scenarios only, where rewards consist of a weighted sum of single-turn compliance and conversion scores.
\end{itemize}

\begin{table}[h]
\centering
\caption{Ablation Study Results.}
\label{tab:ablation}
\begin{tabular}{l|ccc}
\hline
\textbf{Method} & \textbf{CVR (\%)} & \textbf{Compliance($\uparrow$)} & \textbf{Avg. Turn} \\
\hline
\textbf{DuCA (Full)} & \textbf{24.44} & \textbf{66.72} & \textbf{9.91} \\
w/o HIAN & 24.13 & 64.09 & 10.31 \\
w/o Multi-turn & 21.64 & 60.09 & 12.02 \\
\hline
\end{tabular}
\end{table}

The results demonstrate that the multi-turn interaction environment is fundamental to training quality. Without long-term dependency, the model loses business reward guidance, and optimization degenerates to turn-level dominance. Furthermore, without horizon-independent advantage normalization, the gradients from successful conversion cases are diluted by relatively stable turn rewards, leading to unstable convergence.

\subsection{Training Dynamics}

\begin{figure}[htbp]
  \centering
  \begin{subfigure}{0.48\columnwidth}
    \centering
    \includegraphics[width=\linewidth]{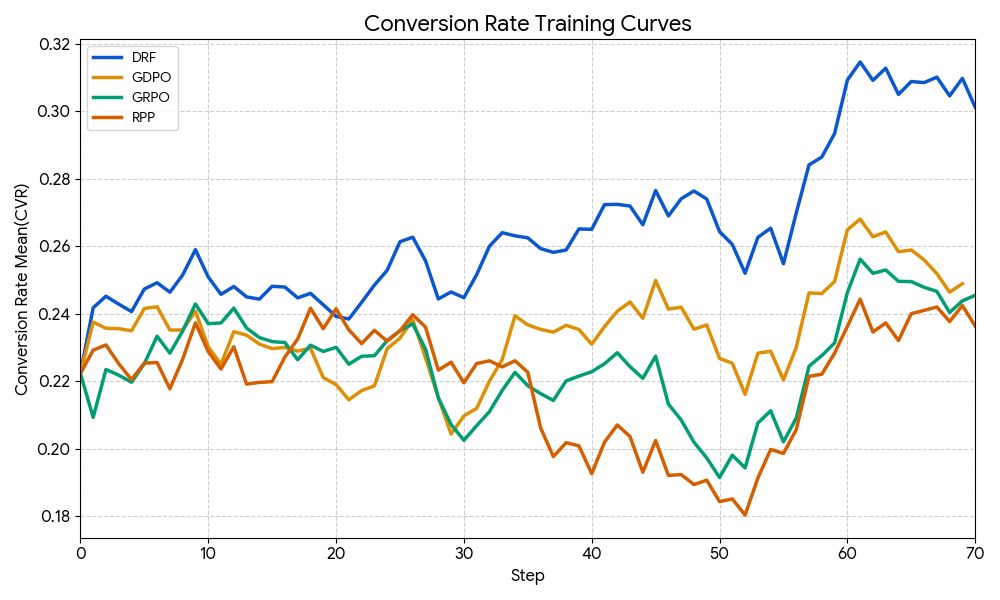}
    \caption{Conversion Score}
    \label{register_curve}
  \end{subfigure}
  \hfill
  \begin{subfigure}{0.48\columnwidth}
    \centering
    \includegraphics[width=\linewidth]{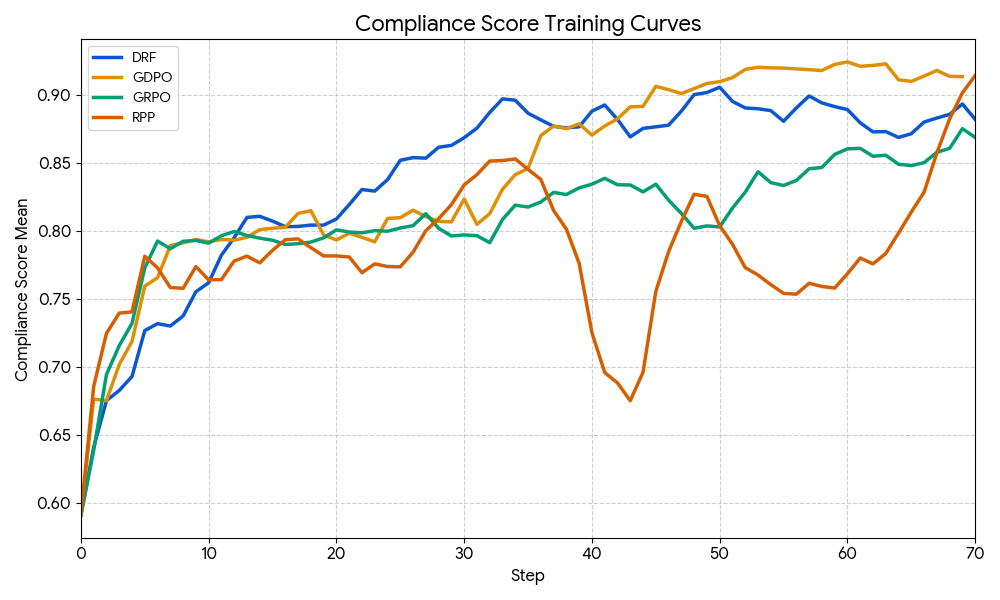}
    \caption{Compliance Score}
    \label{compliance_curve}
  \end{subfigure}
  \caption{Training dynamics of DuCA compared with baselines over 70 steps.}
  \label{fig:curves}
\end{figure}

Figure~\ref{fig:curves} visualizes the optimization stability across all methods. As shown in Figure~\ref{compliance_curve}, REINFORCE++ and GRPO exhibit rapid initial gains in compliance but suffer from significant instability or eventual collapse (e.g., the catastrophic drop in REINFORCE++ after step 35). This illustrates the \textbf{reward hacking trap}: in standard scalarized reward structures, the high-frequency dense turn-level rewards dominate the policy gradient, causing the agent to over-fit to local formatting rules while failing to sustain an upward trend in the sparse, long-term conversion rate.

In contrast, DuCA maintains a remarkably stable and superior learning trajectory in both metrics. By calculating advantages for different temporal horizons independently, our hierarchical decoupling mechanism prevents high-variance session signals from being diluted or overwhelmed by dense rewards. While GDPO achieves slightly higher compliance after step 50, it suffers from a "strategic collapse" in conversion rate due to unconstrained objective competition. DuCA achieves a superior Pareto balance, delivering state-of-the-art conversion performance while strictly adhering to industrial compliance standards.

\subsection{Simulator Fidelity and Feasibility}

To validate the simulator as a reliable proxy for real-world sales exploration, we conduct a multi-dimensional feasibility analysis. The core objective is to ensure the environment is consistent, realistic, and predictive of business outcomes.

We employ GPT-4.1 to evaluate the simulator (v4.7.1) against human expert baselines. As shown in Table \ref{tab:simulator_consistency}, the simulator achieves an average score of \textbf{4.897}, slightly outperforming human experts (4.885). Specifically, the simulator excels in \textit{Tone and Style} (4.969) and \textit{Basic Judgment} (4.910). This indicates that the simulator provides a highly stable and professional conversational environment, which is essential for reducing gradient noise during RL policy updates.

\begin{table}[h]
\centering
\caption{Consistency Evaluation: Simulator vs. Human Level. Scores (1-5) are generated via GPT-4.1 based on persona and context adherence. Best results are \textbf{bolded}.}
\label{tab:simulator_consistency}
\small
\begin{tabular}{lcc}
\hline
\textbf{Metric} & \textbf{Simulator} & \textbf{Human Level} \\ \hline
Receiver Identity & 4.906 & \textbf{4.948} \\
Decision Subject & 4.750 & \textbf{4.824} \\
Merchant Profile & \textbf{4.899} & 4.830 \\
Busy State & 4.957 & \textbf{4.980} \\
Acceptance Level & \textbf{4.888} & 4.830 \\
Basic Judgment & \textbf{4.910} & 4.843 \\
Tone and Style & \textbf{4.969} & 4.941 \\ \hline
\textbf{Average Score} & \textbf{4.897} & 4.885 \\ \hline
\end{tabular}
\end{table}

Beyond linguistic consistency, the simulator demonstrates high behavioral realism. In a blind Turing Test, senior auditors mistakenly identified \textbf{52.0\% of actual human interactions} (detailed in Appendix \ref{appendix: simulator})as being generated by the simulator, suggesting it has captured the professional "gold standard" of sales dialogues. 


\section{Conclusion}

We present DuCA, a robust framework addressing the temporal credit assignment challenge in industrial sales agents. By explicitly decomposing rewards into dual horizons and employing HIAN, our method effectively resolves the gradient dominance and reward hacking issues inherent in multi-granularity optimization. Our work provides a scalable paradigm for aligning LLM agents with complex industrial constraints, ensuring a superior balance between strategic performance and linguistic compliance in real-world deployments.


\section*{Limitations}
\noindent \textbf{Dependency on High-Quality Real-World Data and Human Effort.} The effectiveness of DuCA heavily relies on the high-fidelity user simulator. Constructing such a simulator requires a substantial amount of anonymized real-world business sales dialogues (e.g., 31,000 sessions in our study) and additional high-quality online interaction samples for fine-tuning. This dependency introduces significant human effort in data collection, cleaning, and professional auditing to ensure the simulator captures the "gold standard" of sales interactions. Consequently, deploying DuCA in new industrial domains where such large-scale, domain-specific data or expert human resources are scarce may pose practical challenges for initial model training and simulator calibration.

\section*{Ethical Statement}

\noindent This paper proposes a reinforcement learning framework for industrial sales agents designed to balance strategic performance with compliance. While our approach offers significant benefits in terms of conversion efficiency, it also raises ethical considerations. The use of such agents could lead to unintended consequences, such as bias amplification, where the synthetic agents might inadvertently reinforce existing stereotypes or present skewed sales arguments due to biases in the historical training data.

\noindent Additionally, there is a risk of manipulation of user preferences, as the strategic credit assignment could be used to subtly influence customer behavior without explicit consent. We emphasize that our system incorporates a compliance scoring function to penalize prohibited terms and false promises to mitigate these risks. Furthermore, we suggest that synthetic user simulators should not be a substitute for real human feedback in the long-term design process. Rather, these agents should be leveraged to explore strategies in early stages or high-risk scenarios where direct online exploration with real customers is impractical. By adhering to these principles, we aim to ensure that the deployment of autonomous sales agents is ethical and socially responsible.

\newpage
\bibliography{iclr2026_conference}
\bibliographystyle{colm2025_conference}

\newpage
\appendix
\renewcommand \thepart{} 
    \renewcommand \partname{}
\part{Appendix} 

    \parttoc 
\section{Implementation Details}

\subsection{Reward Function Formulations}
\label{app:reward_fusion}
To ensure the sales agent balances conversational quality with long-term business objectives, we define the reward functions for two distinct temporal granularities.

\subsubsection{Turn-Level Reward ($r_{turn}^{(t)}$)}
\label{app:turn_level}
The turn-level reward provides dense supervision to maintain linguistic constraints and basic dialogue fluency. We employ a \textbf{Gating Fusion} mechanism to prevent the agent from "hacking" simple rewards (e.g., length rewards) by producing repetitive or rigid scripted content. The final reward for turn $t$ is formulated as:

\begin{equation}
r_{turn}^{(t)} = \mathbb{1}_{\text{valid}}(a_t) \cdot r_{\text{len}}(a_t) + (1 - \mathbb{1}_{\text{valid}}(a_t)) \cdot R_{\text{penalty}}
\end{equation}

where the indicator function $\mathbb{1}_{\text{valid}}(a_t)$ is defined by the following gating criteria:

\begin{equation}
\mathbb{1}_{\text{valid}}(a_t) = 
\begin{cases} 
1 & \text{if } \text{Rep}(\tau) \le \delta_1 \text{ or } \text{Sim}(a_t, \mathcal{S}) \le \delta_2 \\
0 & \text{otherwise}
\end{cases}
\end{equation}

The specific components and thresholds are:
\begin{itemize}
    \item \textbf{$\text{Rep}(\tau)$}: Measures the token-level repetition within the current response and the lexical overlap against the previous three dialogue turns. The threshold $\delta_1$ is set to 0.4.
    \item \textbf{$\text{Sim}(a_t, \mathcal{S})$}: Calculates the maximum cosine similarity between the response $a_t$ and the standard professional sales script library $\mathcal{S}$. The threshold $\delta_2$ is set to 0.85 to penalize verbatim template usage without substantive content.
    \item \textbf{$r_{\text{len}}(a_t)$}: A Gaussian-shaped length incentive defined as $r_{\text{len}} = \exp(-\frac{|len(a_t) - L_{\text{target}}|^2}{2\sigma^2})$, where $L_{\text{target}}=30$ tokens.
    \item \textbf{$R_{\text{penalty}}$}: A constant penalty value of -2.0 applied when gating criteria are violated.
\end{itemize}

\subsubsection{Session-Level Reward ($R_{\text{session}}$)}
\label{app:session_level}
The session-level reward (session-level reward) reflects the ultimate industrial goals and is assigned only at the terminal step $T$:

\begin{equation}
R_{\text{session}}(\tau) = \alpha \cdot I(\text{Conversion}) - \beta \cdot S_{\text{violation}}(\tau)
\end{equation}

Where:
\begin{itemize}
    \item \textbf{$I(\text{Conversion})$}: A binary indicator ($1$ for success, $0$ otherwise) determined by an LLM-as-a-Judge evaluating purchase intent.
    \item \textbf{$S_{\text{violation}}(\tau)$}: The cumulative penalty score derived from regulatory audits for prohibited terms or false promises. The \textit{Compliance Score} reported in the main results is the transformed metric $100 - S_{\text{violation}}$.
    \item \textbf{$\alpha, \beta$}: Scaling coefficients used to balance the magnitude of conversion and compliance signals.
\end{itemize}

\subsection{Hyper-parameter and Training Settings}
For the Strategic Credit Assignment via DRF, the optimization parameters are configured as follows:
\begin{itemize}
    \item \textbf{GAE Configuration}: We set $\lambda=1.0$ and $\gamma=1.0$ for both value functions ($V_{\phi_{\text{turn}}}$ and $V_{\phi_{\text{session}}}$). This ensures the session-level reward is propagated back to every turn $t$ without decay, which is critical for long-horizon credit assignment in sales.
    \item \textbf{Fusion Weights}: In our main experiments, we use a balanced ratio of $w_1 = 1.0$ and $w_2 = 1.0$ to ensure both immediate constraints and long-term goals contribute equally to the policy gradient.
    \item \textbf{PPO Setup}: We utilize the PPO algorithm with a KL-divergence penalty term (coefficient 0.05) to constrain policy drift from the SFT initialization.
    \item \textbf{Training Horizon}: All models are trained for a maximum of 70 steps.
\end{itemize}

\subsection{Computing Infrastructure}
Training was conducted on a cluster of 16 $\times$ NVIDIA A100 GPUs (80GB). The sales agent policy $\pi_\theta$ was initialized from a proprietary Large Model pre-trained on 31,000 anonymized real-world business sales dialogues.

\section{Evaluation Reliability and Validity}
\label{appendix: simulator}

To ensure that our reinforcement learning policy is optimized against a representative environment, we evaluate our LLM-based User Simulator across two dimensions: linguistic realism and behavioral consistency.

\subsection{Turing Test Evaluation}
We conducted a blind Turing Test to assess the indistinguishability of our simulator compared to real-world human customers. Three senior sales experts were presented with 250 dialogue snippets (half human-human, half agent-simulator) and asked to identify whether the "customer" was a human or the simulator. 

Table 3 summarizes the experts' judgment on actual human trajectories. Notably, only 36.0\% of human trajectories were correctly identified, while 52.0\% of human interactions were mistakenly classified as being generated by the simulator. This "inverse" confusion suggests that the simulator has captured the core characteristics of professional sales interactions so effectively that it defines the "standard" behavior in the eyes of the auditors.

\begin{table}[h]
\centering
\caption{Human-vs-Simulator Turing Test Results. Results indicate the classification of actual human trajectories by experts.}
\begin{tabular}{lc}
\hline
\textbf{Metric} & \textbf{Percentage (\%)} \\ \hline
Correctly identified as Human & 36.0\% \\
Indistinguishable / Hard to distinguish & 12.0\% \\
Mistakenly identified as Simulator & 52.0\% \\ \hline
\end{tabular}
\end{table}

\subsection{Simulation-to-Reality (Sim-to-Real) Consistency}
A high-fidelity simulator must not only sound realistic but also provide a reliable signal for business outcomes. We compared the Conversion Rates (CVR) obtained in our simulated environment against the actual performance observed in online outbound call scenarios.

We calculated the \textbf{Pearson Correlation Coefficient} across 50 different product categories and persona initializations. The correlation between the simulator's predicted conversion intent and the real-world conversion rate reached:
\begin{equation}
\rho = 0.9733
\end{equation}
This exceptionally high correlation ($\rho > 0.97$) validates that our simulator serves as a robust proxy for real-world customer decision-making processes. Consequently, policy improvements observed in the simulator are highly likely to translate into tangible business gains in production.

\subsection{Persona Adherence}
The simulator's ability to maintain a consistent persona (e.g., \textit{Price-Sensitive, Skeptical}) is monitored via an internal LLM-based auditor. The auditor checks the consistency between the assigned persona prompts and the simulator's responses throughout the multi-turn interaction. Our version 4.6.1 maintains a high degree of adherence, ensuring that the policy is exposed to diverse and stable customer behaviors during training.

\section{Ablation Study Details}

In this section, we provide a more granular analysis of the ablation experiments to further elucidate the impact of the Decoupled Advantage Normalization and the Multi-turn training environment on both business objectives and conversational quality.

\subsection{Detailed Conversational Metrics}
Table \ref{tab:ablation_detailed} presents the fine-grained metrics for the ablation variants. This table complements the high-level results in the main text by showing how each component influences the microscopic behavior of the agent.

\begin{table*}[t]
\centering
\caption{Detailed Ablation Study on Conversational Quality. "w/o Decoupled Norm" refers to the variant using standard reward summation (rpp\_baseline). "w/o Multi-turn" denotes training on single-turn scenarios. Intra-R and Inter-R: Intra-turn and Inter-turn Repetition; IDR: Identity Detection Rate; PATR: Positive Attitude Transfer Rate. Best results are \textbf{bolded}.}
\label{tab:ablation_detailed}
\small
\begin{tabular}{lcccccc}
\hline
\textbf{Variant} & \textbf{Intra-R↓} & \textbf{Inter-R↓} & \textbf{IDR↓} & \textbf{Prefix-R↓} & \textbf{Filler↓} & \textbf{PATR↑} \\ \hline
\textbf{DRF (Full)} & 4.20\% & 2.71\% & \textbf{6.11\%} & \textbf{34.44\%} & 49.21\% & \textbf{44.61\%} \\
w/o Decoupled Norm & \textbf{1.87\%} & \textbf{1.68\%} & 7.35\% & 36.64\% & 52.49\% & 43.83\% \\
w/o Multi-turn & 8.94\% & 41.75\% & 10.34\% & 57.41\% & \textbf{45.71\%} & 43.84\% \\ \hline
\end{tabular}
\end{table*}

\subsection{Training Dynamics and Stability}
\begin{figure}[h]
    \centering
    \includegraphics[width=0.48\textwidth]{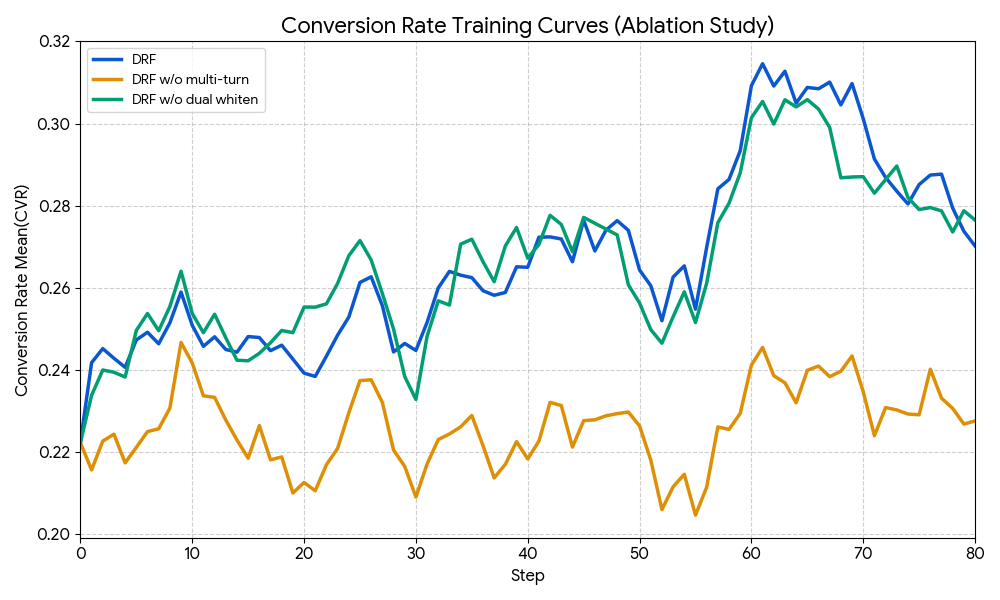}
    \includegraphics[width=0.48\textwidth]{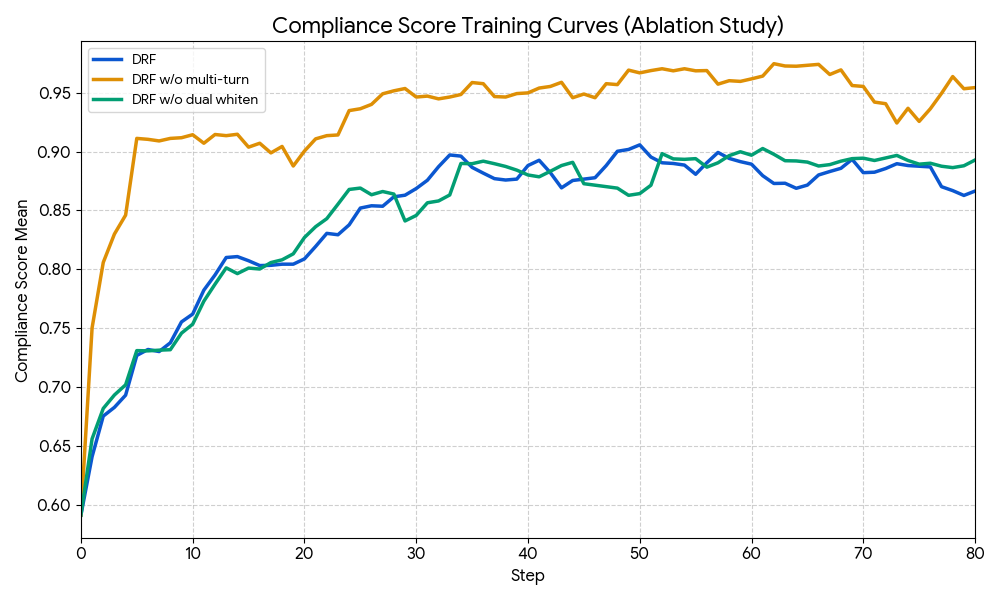}
    \caption{Training dynamics of DRF compared with ablation variants. (a) Conversion Rate (CVR) evolution. (b) Compliance Score evolution.}
    \label{fig:ablation_curves}
\end{figure}

To visualize the optimization process, Figure \ref{fig:ablation_curves} illustrates the evolution of Conversion Rate (CVR) and Compliance scores over the 80-step training trajectory for all variants.

\subsection{Impact Analysis}

\subsubsection{Effectiveness of Decoupled Normalization (Dual Whiten)}
The comparison between \textbf{DRF (Full)} and \textbf{w/o Decoupled Norm} reveals a critical trade-off in multi-objective sales optimization. As shown in Figure \ref{fig:ablation_curves}(a), while the variant without decoupled normalization initially tracks with the full model, it exhibits higher variance and fails to sustain its peak CVR after step 60. 

Furthermore, Table \ref{tab:ablation_detailed} shows that while this variant achieves the lowest repetition rates, its \textbf{Identity Detection Rate (IDR)} is significantly higher (7.35\%) and its \textbf{Positive Attitude Transfer Rate (PATR)} is lower (43.83\%) than the full model. This suggests that without independent advantage normalization, high-variance session-level signals are diluted by stable turn-level rewards, leading the model to converge toward a "safe" but rigid policy that produces robotic responses, ultimately failing to effectively persuade users.

\subsubsection{Necessity of Multi-turn Interaction}
The \textbf{w/o Multi-turn} variant serves as a critical baseline, demonstrating the strategic collapse when long-term dependencies are removed. 

\begin{itemize}
    \item \textbf{Pseudo-Compliance}: In Figure \ref{fig:ablation_curves}(b), this variant reaches high compliance scores faster than others. However, Table \ref{tab:ablation_detailed} clarifies that this is a result of "safe" but repetitive behavior, with \textit{Inter-turn Repetition} (Inter-R) surging to \textbf{41.75\%}. 
    \item \textbf{Strategic Stagnation}: The CVR curve for this variant remains significantly lower and stagnant throughout training. Without a multi-turn interaction loop, the agent cannot capture the causal links between its actions and user intent shifts, validating that our framework requires a multi-turn environment to align dense constraints with sparse business objectives.
\end{itemize}

\end{document}